\documentclass[letterpaper, 10 pt, conference]{ieeeconf}  

\IEEEoverridecommandlockouts                              

\usepackage{xcolor}

\usepackage{amsmath} 
\usepackage{amssymb}  
\usepackage{mathtools}
\usepackage{multicol}
\usepackage{multirow}
\usepackage{booktabs}
\usepackage{pifont}

\title{\LARGE \bf
CRAB: Camera-Radar Fusion for Reducing Depth Ambiguity in\\Backward Projection based View Transformation
}

\author{In-Jae Lee$^{1}$, Sihwan Hwang$^{2}$, Youngseok Kim$^{3}$, Wonjune Kim$^{4}$, Sanmin Kim$^{5}$, and Dongsuk Kum$^{2}$
\thanks{$^{1}$Interdisciplinary Program in Artificial intelligence, Seoul National University. {\footnotesize injae.lee@snu.ac.kr}}
\thanks{$^{2}$ Cho Chun Shik Graduate School of Mobility, KAIST, {\footnotesize [sihhwang0129, dskum]@kaist.ac.kr}}
\thanks{$^{3}$ 42dot Inc, {\footnotesize youngseok.kim@42dot.ai}}
\thanks{$^{4}$ Electronics and Telecommunications Research Institute (ETRI), {\footnotesize wonjune.kim@etri.re.kr}}
\thanks{$^{5}$ Department of Automobile and IT Convergence, Kookmin University, {\footnotesize sanmin.kim@kookmin.ac.kr}}
}

\usepackage[compress]{cite}
\begin{document}

\maketitle
\thispagestyle{empty}
\pagestyle{empty}

\begin{abstract}

Recently, camera-radar fusion-based 3D object detection methods in bird's eye view (BEV) have gained attention due to the complementary characteristics and cost-effectiveness of these sensors. Previous approaches using forward projection struggle with sparse BEV feature generation, while those employing backward projection overlook depth ambiguity, leading to false positives. In this paper, to address the aforementioned limitations, we propose a novel camera-radar fusion-based 3D object detection and segmentation model named CRAB (Camera-Radar fusion for reducing depth Ambiguity in Backward projection-based view transformation), using a backward projection that leverages radar to mitigate depth ambiguity. During the view transformation, CRAB aggregates perspective view image context features into BEV queries. It improves depth distinction among queries along the same ray by combining the dense but unreliable depth distribution from images with the sparse yet precise depth information from radar occupancy. We further introduce spatial cross-attention with a feature map containing radar context information to enhance the comprehension of the 3D scene. When evaluated on the nuScenes open dataset, our proposed approach achieves a state-of-the-art performance among \textit{backward projection-based} camera-radar fusion methods with 62.4\% NDS and 54.0\% mAP in 3D object detection. 
\end{abstract}

\section{INTRODUCTION}

Accurate perception of 3D surroundings plays a crucial role in the fields of autonomous driving and mobile robotics. In autonomous driving and mobile robotics, camera, LiDAR, and radar are the most commonly used sensors for 3D object detection. Due to their different characteristics, the need for research on multi-sensor fusion is increasing. Although research on camera-LiDAR fusion~\cite{supfusion, bevfusion, cmt, autoalignv2} is active, they are susceptible to performance degradation in adverse weather conditions, and the high cost of LiDAR poses a significant obstacle to mass-producing autonomous vehicles. Recently, attention has shifted towards research that fuses low-cost camera and radar, thanks to their complementary characteristics. Specifically, 
images captured by the camera contain dense semantic and contextual information in pixel form along the angular direction.
However, 
the cameras cannot accurately determine the 3D spatial distance, referred to as depth, to objects as it is an ill-posed  problem. In contrast, radar provides accurate distance information and is robust in adverse weather conditions~\cite{grifnet}. Despite their advantages, radar point clouds suffer from limitations, including sparsity, noise from multi-path effects, and limited angular resolution~\cite{craft}. Therefore, by optimally fusing these complementary sensors, we can leverage their strengths and minimize their limitations, thereby significantly enhancing performance.
\begin{figure}[t]
    \begin{center}
    \vspace{-5pt}
\includegraphics[width=1\linewidth]{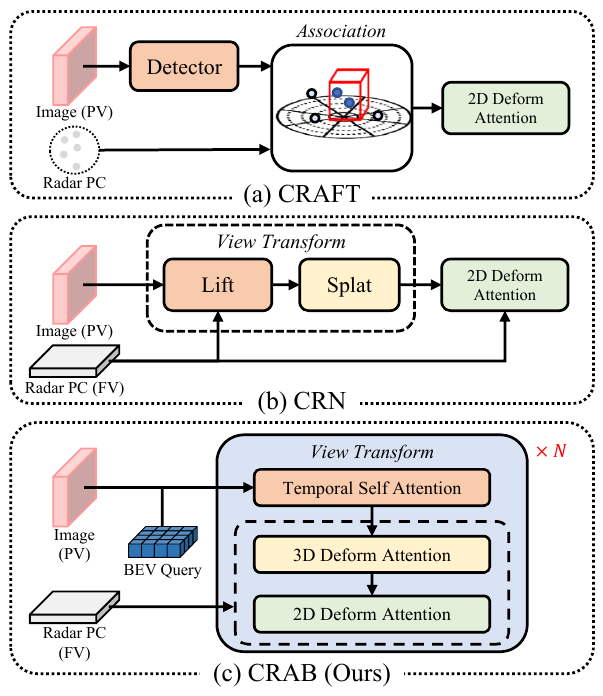}
    \end{center}
    \caption{\textbf{Comparison of camera-radar fusion methods.} (a) In CRAFT, 3D proposals are generated directly from the perspective view image using a detector, and fusion involves filtering noisy radar points outside the proposals and 2D deformable attention. (b) CRN utilizes forward projection, incorporating radar occupancy in the `lift' stage, and adaptively fuses image and radar BEV features with 2D deformable attention without refinement. (c) CRAB, based on backward projection, employs radar occupancy across multiple layers for 3D deformable attention, ensuring clear depth distinction while transforming image context to BEV. Subsequently, 2D deformable attention aggregates radar context features into the BEV query.}
    \label{fig1}
\end{figure}

 Previous fusion approaches~\cite{craft,centerfusion} typically utilize 3D object proposals obtained from perspective view images, followed by filtering noisy radar point clouds, as shown in Fig.~\ref{fig1} (a). Fusion then occurs through operations like attention~\cite{craft} or concatenation~\cite{centerfusion}. However, these methods heavily rely on the performance of camera-only 3D object detection model and may not be suitable for downstream tasks such as segmentation, path planning and prediction which occur in BEV space, leading to space discrepancies. Meanwhile, 
recent research on camera-radar fusion~\cite{crn,rcmfusion,rcbev,transcar,redformer} is predominantly conducted in the BEV space due to advantages such as reduced variance in object sizes, agnostic to other tasks and integration of coordinate systems from various sensors. These studies primarily leverage two existing methods of view transformation: forward projection and backward projection. Approaches based on forward projection~\cite{crn},~\cite{rcbev} fail to address issues such as the generation of sparse BEV features as the distance from the ego vehicle increases and heavy dependence on depth prediction due to the absence of refinement processes~\cite{fb-bev}. On the other hand, research utilizing backward projection~\cite{rcmfusion,redformer} inadequately tackles the issue of BEV queries along the same ray obtaining the same image features due to the absence of depth distinction.

Therefore, to overcome the aforementioned limitations, we propose CRAB, a backward projection~\cite{bevformer} based 3D object detection and segmentation model, leveraging camera-radar fusion to mitigate the depth ambiguity. Our first proposed module, Radar Occupancy-guided Spatial Cross Attention (ROSCA), generates accurate BEV features by incorporating radar occupancy for more precise depth estimation, in addition to the depth distribution used in the previous DFA3D~\cite{dfa3d} method. Additionally, to leverage radar context information, we perform deformable attention~\cite{deformable} between the frustum view-shaped radar context feature map and the BEV query in Radar Context-aware Spatial Cross Attention (RCSCA). In RCSCA, even if queries are on the same ray, they are projected to different depth positions, allowing us to obtain depth-distinguishable features. CRAB achieves the best performance among camera-radar fusion models based on the same view transformation (backward projection) method when tested on the nuScenes~\cite{nuscenes} open dataset. Furthermore, additional experiments validate the effectiveness of our proposed approach. Our main contributions are as follows:
\begin{itemize}
\item{
We alleviate the depth ambiguity issue in the backward projection-based view transformation method through camera-radar fusion. By proposing two modules (ROSCA, RCSCA), we generate semantically rich and spatially accurate BEV features that contain the understanding of the 3D scene.
}
\item{When conducting experiments on the nuScenes open dataset, our proposed architecture, CRAB, exhibits the best performance among \textit{backward projection-based methods} in 3D object detection camera-radar track.}

\end{itemize}

\section{Related Work}
\label{sec:Related Work}
There exists a view discrepancy between the images and the radar point cloud. Therefore, it is necessary to fuse data obtained from these two different sensors in a unified space.

\begin{figure*}[t]
    \centering
    \includegraphics[width=1\textwidth]{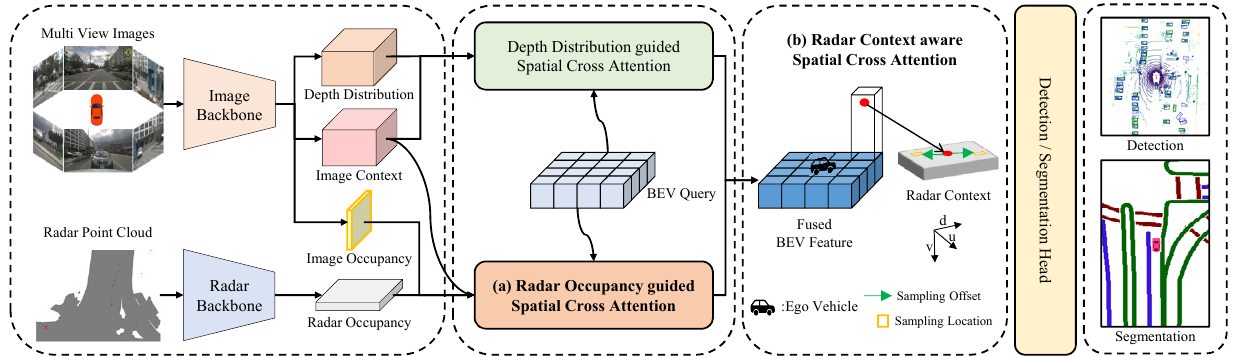}
    \vspace{-5pt}
    \caption{ \textbf{Overall architecture of CRAB}. CRAB extracts image and radar features from their respective backbones. Afterwards, BEV features obtained through Depth Distribution guided Spatial Cross Attention and (a) Radar Occupancy guided Spatial Cross Attention are fused together. The fused BEV feature and radar context features undergo (b)spatial cross attention, before finally passing through a task-specific heads.
    }
    \label{fig2}
    \vspace{-15pt}
\end{figure*}

\subsection{Fusion in Perspective View}
GRIF Net~\cite{grifnet} fuses the Region of Interest (RoI) obtained from each modality through a gating mechanism. Centerfusion~\cite{centerfusion} and CRAFT~\cite{craft} associate 3D proposals obtained from the perspective view with noisy radar points to filter. While Centerfusion~\cite{centerfusion} uses concatenation along the channel direction, CRAFT~\cite{craft} employs deformable attention~\cite{deformable} for fusion. MVFusion~\cite{mvfusion} aligns semantic information from radar and image features, then performs fusion in the global field using cross-attention. However, those approaches are not easily extended to other downstream tasks.

\subsection{Fusion in Bird's Eye View}
 Thanks to significant performance growth in camera-based 3D object detection in BEV, subsequent studies such as~\cite{crn,rcmfusion,rcbev,transcar,redformer} perform fusion in a BEV unified space. Both~\cite{crn} and~\cite{rcbev} are based on forward projection view transform. CRN~\cite{crn} utilizes radar occupancy during the `lift' process and resolves spatial misalignment by adaptively aggregating features from image and radar BEV features through deformable attention~\cite{deformable}. RCBEV~\cite{rcbev} concatenates BEV features from different modalities. Then, it fuses the obtained image feature map's heatmap and radar heatmap via CNN. However, neither approach addresses the inherent issue of sparse BEV generation in forward projection. RCMFusion~\cite{rcmfusion} and TransCAR~\cite{transcar} both rely on backward projection. \cite{rcmfusion} incorporates radar guidance at the feature level to encode the BEV, followed by refinement at the instance level.~\cite{transcar} utilizes soft-association between vision-updated queries and radar features in the decoder instead of hard-association based on sensor calibration. However, both methods still fail to address the critical issue of depth ambiguity during the view transformation process.

\section{Methodology}

CRAB consists of two modules that mitigate the depth ambiguity issues in previous approaches~\cite{rcmfusion,redformer,bevformer},  and effectively fuse different modalities. As illustrated in Fig.~\ref{fig2}, our model comprises image and radar backbones, Depth Distribution \& Radar Occupancy-guided Spatial Cross Attention, Radar Context-aware Spatial Cross Attention, and task specific heads. The Depth Distribution \& Radar Occupancy-guided Spatial Cross Attention module accurately transforms the perspective view image context features into the BEV space using depth information from depth distribution and radar occupancy. Then, the subsequent Radar Context-aware Spatial Cross Attention module integrates radar context information into the features. Finally, the heads perform downstream tasks, such as detection and segmentation, on BEV features to produce the final results.

\subsection{Preliminaries}
\subsubsection{BEVFormer} 
BEVFormer~\cite{bevformer} is a camera-based 3D object detection model using backward projection-based view transformation. For view transformation, points $\mathbf{x}=(x,y,z)$ distributed along the z-axis in  BEV query $\mathbf{Q}\in\mathbb{R}^{C\times X\times Y}$ are projected onto N perspective view image context features $\mathbf{I}_{C} \in\mathbb{R}^{N \times C\times H\times W}$  using camera parameters $\mathbf{P}$. Here, $C, X, Y, H$, and $\mathit{W}$ denote channel, size of BEV grids and image resolution, respectively. Features that need to be sampled, based on the projected points, are learned through deformable attention~\cite{deformable}. Then BEV query is updated as $\mathbf{Q(x)}'$  with aggregation of weighted sampled features as follows:
\vspace{-3pt}
\begin{equation}
\label{equation:BEVFormer}
\mathbf{Q(x)'} = DeformAttn(\mathbf{I}_{C},\mathbf{Q(x)},\mathbf{P}) .
\end{equation}
\subsubsection{DFA3D} Due to the camera geometry, in BEVFormer~\cite{bevformer}, 3D points on BEV queries along the same ray are projected onto the image feature map at the same location, leading to the drawback of obtaining the same features without distinguishing depth. To address this limitation, DFA3D~\cite{dfa3d} takes lifted $\mathbf{I}_{CD}\in\mathbb{R}^{N\times C\times D\times H\times W}$ as value instead, which is the result of the outer product between depth distribution $\mathbf{I}_{D}\in\mathbb{R}^{N\times D\times H\times W}$ and image context feature $\mathbf{I}_{C}\in\mathbb{R}^{N\times C\times H\times W}$. This enables the queries along the same ray to be projected onto different points, allowing the retrieval of features with distinct depth values.

\subsection{Image \& Radar Point cloud Processing}
For the camera stream, we adopt the approach of BEVFormer-DFA3D~\cite{dfa3d}. Given six images covering a 360-degree field of view (FoV) around the vehicle, we use an image backbone (\textit{e.g.}, ResNet~\cite{resnet}, V-99~\cite{vov}), and FPN~\cite{fpn} to extract multi-scale image feature maps. Then these image feature maps pass through DepthNet~\cite{depthnet} following the methodologies of BEVDepth~\cite{bevdepth} and LSS~\cite{lss}, resulting in obtaining a discrete depth distribution $\mathbf{I}_{D}\in\mathbb{R}^{N\times D\times H\times W}$. Additionally, employing multiple CNN layers and softmax function, we obtain $\mathbf{I}_{O}\in\mathbb{R}^{N\times 1\times H\times W}$ containing occupancy information and $\mathbf{I}_{C}\in\mathbb{R}^{N\times C\times H\times W}$ containing context information. \par
On the other hand, for the radar stream, a point cloud with coordinates in the radar coordinate system is projected on perspective view images to find corresponding pixels and voxelized into camera frustum view voxels following CRN~\cite{crn}. During this projection, the depth in the camera-based coordinate system is preserved, resulting in coordinates in a pillar-like~\cite{pointpillar} frustum view (u,v=1,d). In this view, u,v denotes width and height in the pixel coordinate system respectively and d denotes depth in the camera coordinate system. Subsequently, after passing through PointNet~\cite{pointnet} and SECOND~\cite{second}, we extract the frustum view-shaped features $\mathbf{R}_{O}\in\mathbb{R}^{N\times D\times 1\times W}$ containing occupancy information and $\mathbf{R}_{C}\in\mathbb{R}^{N\times C\times D\times W}$ containing radar context information such as Radar Cross Section (RCS) and the Doppler velocity. 

\subsection{View Transformation}
\subsubsection{Radar Occupancy-guided Spatial Cross Attention}
\label{ssec:ROSCA}
In DFA3D~\cite{dfa3d}, to address the same-ray-same-feature problem, it adopts depth-weighted bilinear interpolation operations. These operations are based on probability information within the predicted depth distribution  $\mathbf{I}_{D}\in\mathbb{R}^{N\times D\times H\times W}$ from the image. However, it is difficult to distinguish image context features accurately due to the inherently inaccurate depth information obtained from monocular depth estimation~\cite{depthnet}.
Therefore, we propose to use sparse yet accurate radar occupancy, similar to~\cite{crn}, to better distinguish the depth information of image context features.
As shown in Fig.~\ref{fig3}, we obtain 3D occupancy denoted as $\mathbf{O}_{IR}\in\mathbb{R}^{N\times D\times H\times W}$, by taking the outer product of the perspective view image occupancy $\mathbf{I}_{O}\in\mathbb{R}^{N\times 1\times H\times W}$ and the frustum view radar occupancy $\mathbf{R}_{O}\in\mathbb{R}^{N\times D\times 1\times W}$. Next, by taking the outer product with image context $\mathbf{I}{c}\in\mathbb{R}^{N\times C\times H\times W}$, we obtain an extended feature map $\mathbf{X}\in\mathbb{R}^{N\times C\times D\times H\times W}$. Now, through depth-aware spatial cross attention operations (DFA3D)~\cite{dfa3d} between the extended feature map $\mathbf{X}\in\mathbb{R}^{N\times C\times D\times H\times W}$ and the BEV query $\mathbf{B}\in\mathbb{R}^{C\times X\times Y}$, we obtain $\mathbf{B}_{R}\in\mathbb{R}^{C\times X\times Y}$ which encodes depth-aware image features. The following equation represents the whole process:

\begin{figure}[t]
    \begin{center}
\includegraphics[width=0.8\linewidth]{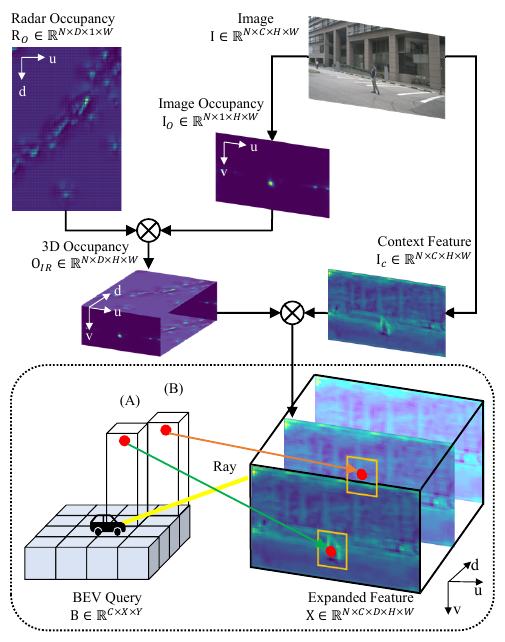}
    \end{center}
    \vspace{-15pt}
    \caption{\textbf{Illustration of proposed Radar Occupancy-guided Spatial Cross Attention.} The 3D occupancy generated by the outer product of radar occupancy and image occupancy, along with the outer product of image context feature, is used to obtain an expanded feature. Through spatial cross-attention operations between the BEV query and the expanded feature, features are aggregated into the BEV query. Even though points (A) and (B) are on the same ray, they are projected on the different depths, allowing us to encode depth distinguished features.}
    \label{fig3}
\end{figure}

\begin{equation}
\begin{split}
    & \mathbf{O}_{IR}= \mathbf{I}_{O} \otimes \mathbf{R}_{O} , \mathbf{X}= \mathbf{O}_{IR} \otimes \mathbf{I}_{C}    \\ 
    & \mathbf{B}_{R}=\textit{DeformAttn3D}(\mathbf{X}, \mathbf{B},\mathbf{P}). 
\label{eq:ouccpancy guided bevfeature}
\end{split}
\end{equation}
In addition to the $\mathbf{B}_{R}$, we also obtain image context encoded BEV feature $\mathbf{B}_{I}\in\mathbb{R}^{C\times X\times Y}$ utilizing the depth distribution~\cite{dfa3d}.
We fuse these two BEV features into a single BEV feature $\mathbf{B}_{IR}$ through the following equation:
\begin{equation}
\begin{split}
    &\mathbf{F}_{I} =  \psi(\mathcal{M} (\mathbf{B}_{I})) , \mathbf{F}_{R} =  \psi(\mathcal{M} (\mathbf{B}_{R})) \\
    & \mathbf{B}_{IR}= \psi^{-1}(\xi((\mathbf{F}_{I};\mathbf{F}_{R}))),
\label{eq:fusion}
\end{split}
\end{equation}
where $\mathcal{M}$ denotes mask filtering the projected points which are outside of feature map, $\psi$ is the flatten operation, ; is a concatentation along channel dimension and $\xi$ denotes MLP network.

\subsubsection{Radar Context-aware Spatial Cross Attention}
In the previously introduced module (ROSCA), the radar is utilized to provide occupancy in the frustum view to incorporate image context features better into BEV features.
In addition to occupancy, radar point cloud provides Radar Cross Section (RCS) and the Doppler velocity $\mathit{V}_{x},\mathit{V}_{y}$.\begin{table*}[t]
\centering
\caption[3D Object Detection on nuScenes \texttt{val} set. ]
{
     Quantitative Results of 3D Object Detection on nuScenes \textit{validation} set. `L', `C', and `R' represent LiDAR, camera, and radar respectively.
    $^*$: results from MMDetection3D~\cite{mmdet3d}.
    $^\dagger$ is initialized from FCOS3D backbone.
    $^\ddagger$ is trained with CBGS. \textbf{Bold} and \underline{underline} denote the best value and second best results among C and C\&R. V.T is View Transformation.
}
\begin{center}
\begin{tabular}{l|c|c|c||cc|ccccc}
\hline
Method        & Modality & Backbone & V.T Method  & NDS$\uparrow$           & mAP$\uparrow$           & mATE$\downarrow$           & mASE$\downarrow$           & mAOE$\downarrow$           & mAVE$\downarrow$           & mAAE$\downarrow$           \\ \hline \hline
CenterPoint-P$^{\dagger*}$\cite{centerpoint} & L        & Pillars   & -         & 59.8          & 49.4          & 0.320          & 0.262          & 0.377          & 0.334          & 0.198          \\
CenterPoint-V$^{\dagger*}$\cite{centerpoint} & L        & Voxel        & -      & 65.3          & 56.9          & 0.285          & 0.253          & 0.323          & 0.272          & 0.186          \\ \hline
MatrixVT\cite{matrixvt} & C        & V2-99  & Forward     & 56.2          & 46.6          & 0.535          & $\underline{0.260}$          & 0.380          & 0.342          & 0.198\\  
BEVFormer-S-DFA3D\cite{dfa3d}   & C        & ResNet101  & Backward     & 50.1          & 40.1          & 0.721         & 0.279          & 0.411         & 0.391          & 0.196          \\
BEVFormer-DFA3D\cite{dfa3d}   & C        & ResNet101  & Backward   & 53.1          & 43.0          & 0.654         & 0.271          & 0.374         & 0.341          & 0.205          \\
REDFormer\cite{redformer}     & C\&R      & ResNet101  & Backward     & 48.6          & 38.5          & 0.726          & 0.282          & 0.407          & 0.427          & 0.218          \\ 
TransCAR\cite{transcar}      & C\&R      & ResNet101  & Backward         & 49.7          & 38.1          & 0.526         & 0.272  & 0.445         & 0.465          & 0.185          \\
RCM-Fusion$^{\ddagger}$\cite{rcmfusion}    & C\&R      & ResNet101  & Backward     & 58.3          & 49.7          & 0.508              & 0.260              & \textbf{0.325}              & 0.377              & 0.181              \\\hline
CenterFusion\cite{centerfusion}  & C\&R      & DLA & -         & 45.3          & 33.2          & 0.649          & 0.263          & 0.535          & 0.540          & \textbf{0.142} \\
CRAFT \cite{craft}  & C\&R      & DLA34   & -       & 51.7          & 41.1          & 0.494          & 0.276          & 0.454         & 0.486          & 0.176 \\
MVFusion$^{\ddagger}$\cite{mvfusion}      & C\&R      & ResNet101 & -            & 45.5          & 38.0          & 0.675          & \textbf{0.258} & 0.372          & 0.833          & 0.196          \\
RCBEV4d$^\ddagger$\cite{rcbev}      & C\&R      & Swin   & Forward         & 49.7          & 38.1          & 0.526         & 0.272  & 0.445         & 0.465          & 0.185          \\
CRN\cite{crn}           & C\&R      & ResNet101 & Forward      & 59.2         & \underline{52.5}          & \textbf{0.460} & 0.273          & 0.443          & 0.352          & 0.180          \\
\textbf{CRAB}          & C\&R      & ResNet101  & Backward      & $\underline{59.7}$& 51.7& 0.506          & 0.267          & 0.358 & \underline{0.300} & 0.185          \\
\textbf{CRAB}          & C\&R      & V2-99  & Backward     & \textbf{62.4} & \textbf{54.4} & $\underline{0.503}$          & 0.266          & $\underline{0.280}$& $\textbf{0.277}$& $\underline{0.170}$          \\ \hline
\end{tabular}
\end{center}
\label{table1}
\end{table*}To leverage this radar context information, we perform spatial cross-attention operations~\cite{bevformer} between the radar context feature map and the previously image context encoded BEV feature.

In this case, the 3D points from the BEV feature are projected onto the frustum-shaped radar context feature, which has a height of 1. As observed in Fig.~\ref{fig2} (b), different points on the same ray can retrieve distinct features, since the radar features are extracted along the $(u,d)$ direction. 
\begin{figure}[t]
    \begin{center}
\includegraphics[width=1\linewidth]{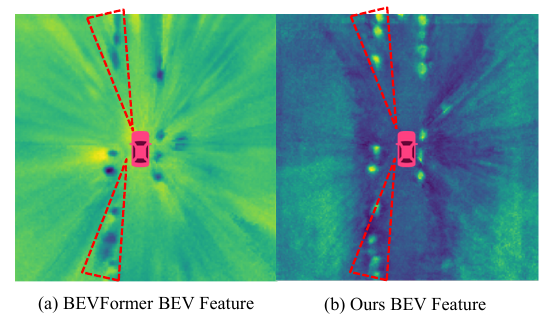}
    \end{center}
    \vspace{-15pt}
    \caption{\textbf{Comparison of Encoded BEV feature.} During view transformation, (a) does not consider depth, resulting features in red triangular area(ray area) being somewhat scattered and indistinct. In contrast, the BEV feature of the proposed architecture (b) shows a clear distinction in the depth of features, demonstrating improved clarity.}
    \label{fig4}
\end{figure}
Furthermore, by using a frustum view shaped rather than BEV-shaped radar context feature, CRAB can consistently learn without view discrepancies during the spatial cross-attention operation~\cite{bevformer}, enabling it to find correspondence with image features effectively. Table~\ref{table5} shows the effectiveness of our design choice for the frustum view-shaped radar context feature.
The process can be represented by the equation below: 
\begin{equation}
\begin{split}
    & \mathbf{B}_{encoded}=\textit{DeformAttn}
    (\mathbf{R}_{C}(u,d),\mathbf{B}_{IR},\mathbf{~\tilde{P}}),
\label{eq:image feature extract}
\end{split}
\end{equation}
where $\mathbf{~\tilde{P}}$ indicates the projection matrix from 3D points on BEV to the frustum view radar context feature map. After passing through the two modules presented earlier, as shown in Fig.~\ref{fig4} (b), we finally obtain BEV features where both image and radar context features are encoded. These BEV features then pass through decoders tailored to each downstream task, such as 3D object detection and segmentation. 

\subsection{Loss Function}

To train CRAB, three types of losses are required: detection loss $\mathcal{L}_{detection}$, depth loss $\mathcal{L}_{depth}$, and occupancy loss $\mathcal{L}_{occupancy}$, as formulated below:
\begin{equation}
\label{equation:Loss}
\mathcal{L}_{total} = \mathcal{L}_{detection}+\mathcal{L}_{depth}+\mathcal{L}_{occupancy}.
\end{equation}
The detection loss, denoted as $\mathcal{L}_{detection}$, includes the L1 loss for bounding box regression and the focal loss for classification. We employ binary cross-entropy loss for $\mathcal{L}_{depth}$ to supervise the depth distribution. Finally, the image occupancy loss $\mathcal{L}_{occupancy}$ utilizes Gaussian Focal Loss with the heatmap from the 2D bounding box, which is regarded as the ground truth and is obtained from the projected 3D bounding box annotations.

\section{Experiments}
\subsection{Datasets and Metrics}
We evaluate our proposed architecture on the nuScenes open dataset ~\cite{nuscenes}. We use 6 TP metrics and NDS for 3D object detection and mIoU for segmentation. Please refer to ~\cite{bevformer} and ~\cite{nuscenes} for more details about the dataset and metrics. 

\setlength{\tabcolsep}{4pt}
\renewcommand{\arraystretch}{1.0}

\begin{table}[!hbt]
\vspace{-10pt}
\caption[3D Object Detection on nuScenes \texttt{test} set. ]
{Results of 3D Object Detection on nuScenes \textit{test} set. \textbf{Bold} and \underline{underline} denote the best value and second best results.}
\begin{tabular}{l|c|c|ccc}
\hline
Method       & Modality & BackBone   & mAP$\uparrow$  & NDS$\uparrow$  & mAVE$\downarrow$  \\ \hline \hline
PointPillars\cite{pointpillar} & L        & Pillars    & 40.1 & 55.0 &-       \\
CenterPoint\cite{centerpoint}  & L        & Voxel      & \textbf{60.3} & \textbf{67.3} &-       \\ \hline
BEVFormer\cite{bevformer}    & C        & V2-99      & 48.1 & 56.9 & 0.378 \\
BEVDepth\cite{bevdepth}     & C        & ConvNext-B & 52.0 & 60.9 & 0.347 \\ \hline
CenterFusion\cite{centerfusion} & C\&R     & DLA34      & 33.2 & 45.3 & 0.540  \\
CRAFT\cite{craft}        & C\&R     & DLA34      & 41.1 & 52.3 & 0.519 \\
MVFusion\cite{mvfusion}     & C\&R     & V2-99      & 45.3 & 51.7 & 0.781 \\
RCMFusion\cite{rcmfusion}    & C\&R     & ResNet101  & 50.6 & 58.7 & 0.438 \\
RCBEV\cite{rcbev}  & C\&R     & Swin       & 47.6 & 56.7 & 0.439 \\
TransCAR\cite{transcar}     & C\&R     & V2-99      & 42.2 & 52.2 & 0.495 \\
\textbf{CRAB}          & C\&R     & V2-99      & \underline{54.0} & \underline{62.4} & \textbf{0.270} \\ \hline
\end{tabular}
\label{table2}
\vspace{-15pt}
\end{table}

\subsection{Implementation Details}
We adopt BEVFormer-DFA3D~\cite{dfa3d} as a baseline. We utilize the temporal self-attention module and detection head employed in BEVFormer~\cite{bevformer} without modification. For images, we use ResNet101-DCN~\cite{resnet} and VovNet~\cite{vov} as the backbone. Following GRIF Net~\cite{grifnet}, we accumulate 6 previous sweeps of radar point cloud. We use 6 encoder layers, and the BEV grid size is 
200$\times$200 for Table~\ref{table1},~\ref{table2}, and~\ref{table6}. Also we use 3 encoder layers and BEV grid size is 150$\times$150 Table~\ref{table3},~\ref{table4}, and~\ref{table5}. For a fair comparison with the baseline~\cite{dfa3d}, we do not use any data augmentation, future frames, CBGS~\cite{cbgs}, and Test Time Augmentation (TTA). We conduct experiments for 24 epochs with AdamW~\cite{adam} optimizer and  $2 \times 10^{-4}$ learning rate on 4 NVIDIA RTX3090 GPUs.
\subsection{Evaluation Results}

\subsubsection{3D Object Detection}

\begin{figure}[t]
    \centering
    \includegraphics[width=1\linewidth]{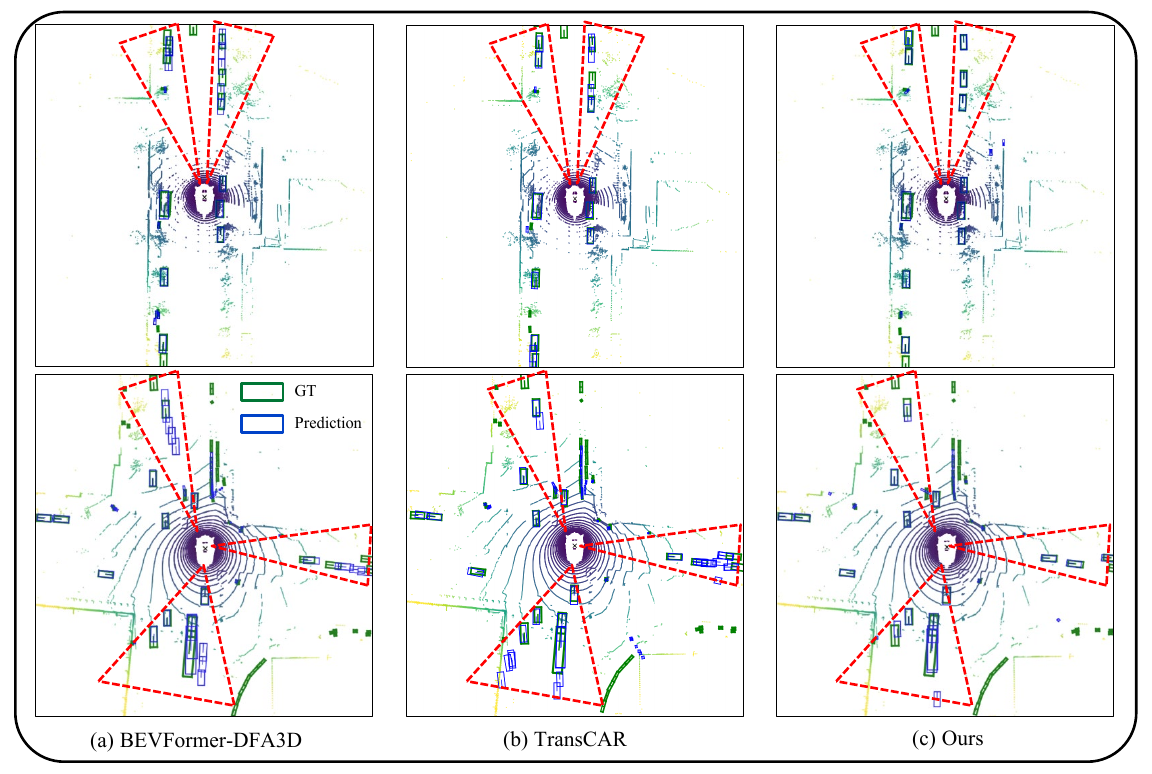}
    \vspace{-15pt}
    \caption{\textbf{Qualitative results of 3D object detection}. When observing the red-ray area, BEVFormer-DFA3D and TransCAR predict multiple instances for a single object, leading to numerous false positives. In contrast, CRAB significantly reduces false positives. Best viewed in color with zoom in. 
    }
    \label{qualitativedetection}
    \vspace{-5pt}
\end{figure}

Table~\ref{table1} and Table~\ref{table2} present the results of experiments conducted on the validation and test sets of nuScenes~\cite{nuscenes}, respectively. CRAB achieves significant improvement compared to camera-only methods~\cite{dfa3d,matrixvt}. Also, our method outperforms baselines (\textit{e.g.},~\cite{bevformer,dfa3d}) which uses camera-only and camera-radar fusion methods~\cite{redformer,rcmfusion}. This suggests that CRAB effectively fuses camera and radar based on backward projection-based compared to other methods such as~\cite{redformer,rcmfusion}. Furthermore, our method shows improvement compared to other fusion methods (\textit{e.g.},~\cite{crn,craft,transcar}), highlighting the potential of radar fusion in a backward projection-based approach. Lastly, the slightly lower or superior performance compared to LiDAR-only methods~\cite{centerpoint} suggests the potential of fusion between camera and radar.\begin{table}[!hbt]
\vspace{-10pt}
\caption[Results of Segmentation on nuScenes data set.]
{Results of Segmentation on nuScenes  dataset. \textbf{Bold} and \underline{underline} denote the best value and second best results. }
\begin{tabular}{l|c ||c c c c}
\hline
Method           & Input & Divider & Ped & Boundary & mIoU$\uparrow$ \\ \hline \hline
LSS\cite{lss} & C     & 38.3    & 14.9         & 39.3     & 30.8 \\
HDMapNet(Surr)\cite{hdmapnet}  & C     & 40.6    & 18.7         & 39.5     & 32.9 \\
BeMapNet\cite{bemapnet}         & C     & 49.1    & $\underline{42.2}$& 39.9     & 43.7 \\
BEVSegFormer\cite{bevsegformer}     & C     & \textbf{51.1}    & 32.6         & 50.1    & 44.6 \\
MapTR+MapVR\cite{mapvr}      & C     & 47.7    & \textbf{54.4}& 51.4     & \textbf{51.2} \\
BEVFormer\cite{bevformer}        & C     & 47.1    & 36.7         & $\underline{52.3}$& 45.6 \\
\textbf{CRAB}             & C\&R   & $\underline{49.5}$& 39.0         & \textbf{57.0}     & $\underline{48.5}$\\ \hline
\end{tabular}
\label{table3}
\vspace{-15pt}
\end{table}Fig.~\ref{qualitativedetection} shows a qualitative comparison between the baseline~\cite{dfa3d}, TransCAR~\cite{transcar} and CRAB. When examining the triangular region corresponding to the ray area centered around the ego vehicle, we observe that, despite utilizing depth distribution, the baseline method predicts multiple instances for a single object (ground truth). Moreover, the occurrence of this phenomenon in TransCAR~\cite{transcar} suggest that, despite using the same view transformation (backward projection), it does not consider depth ambiguity. In contrast, CRAB effectively alleviates depth ambiguity by leveraging accurate depth information from radar, resulting in a noticeable reduction in false positives.

\subsubsection{Segmentation}

\begin{figure}[!hbt]
    \centering
    \includegraphics[width=1\linewidth]{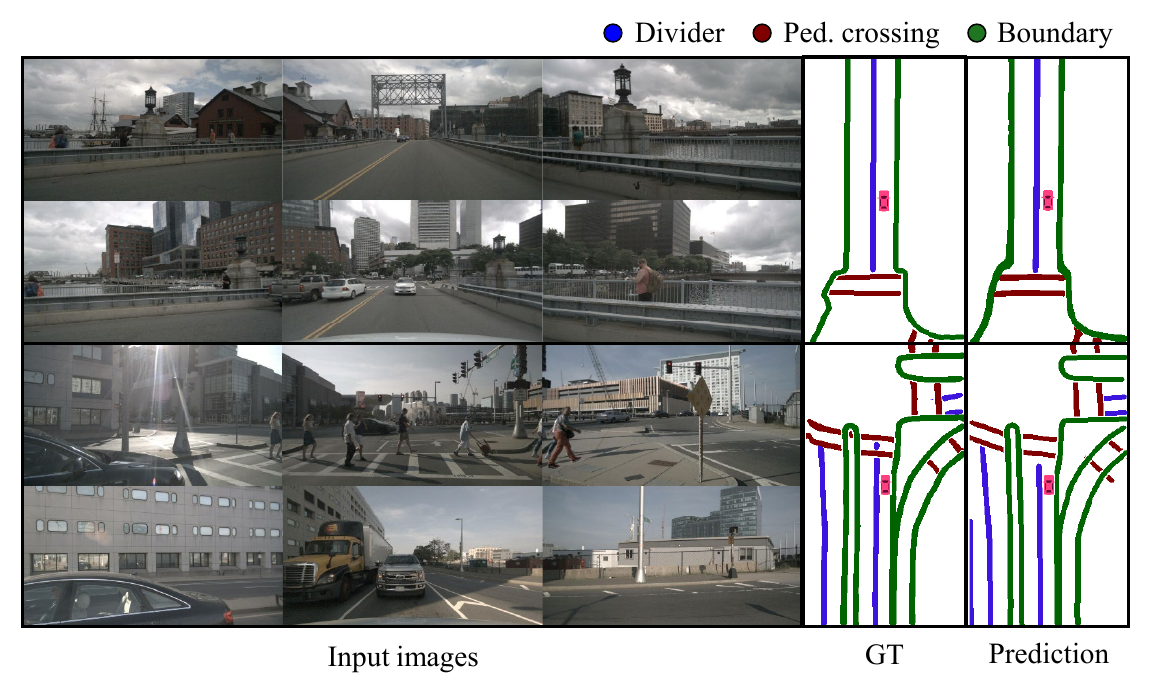}
    \vspace{-15pt}
    \caption{\textbf{Qualitative results of BEV segmentation on nuScenes \textit{val} set.} Our proposed method demonstrates the ability to construct maps accurately in complex scenarios. Best viewed in color with zoom in. 
    }
    \label{segmentation}
    \vspace{-5pt}
\end{figure}

We also conduct segmentation as depicted in Table~\ref{table3} and Fig.~\ref{segmentation}. The segmentation head comprises multiple CNN layers, predicting three classes: divider, pedestrian, and boundary. We use cross-entropy loss for training segmentation. Compared to the BEVFormer~\cite{bevformer}, our method shows improved performance across all classes, with results comparable to other state-of-the-art models. This indicates that CRAB generates general and sophisticated BEV features, enabling it to perform various perception tasks effectively.

\subsection{Ablation studies \& Analysis}

\subsubsection{Component Analysis} We conduct an ablation study to assess the effectiveness of modules\begin{table}[h]
\caption[Ablation Study]
{
    Ablation study of ROSCA \& RCSCA.
}
\begin{center}
\begin{tabular}{c|cc|c|c|c}
\hline
    & ROSCA & RCSCA & NDS$\uparrow$  & mAP$\uparrow$  & mATE$\downarrow$  \\ \hline\hline
(A) &                &               & 50.1 & 40.1 & 0.721 \\    
(B) & $\checkmark$&               & 57.0 & 48.7 & 0.576 \\
(C) &                 & $\checkmark$& 58.0 & 49.4 & 0.542 \\
(D) & $\checkmark$& $\checkmark$& \textbf{58.5} & \textbf{49.9} & \textbf{0.527} \\ \hline
\end{tabular}
\label{table4}
\end{center}
\end{table}\begin{table}[hbt!]
\vspace{-10pt}
\caption
{
    Analysis of Radar Context Feature view.
}
\begin{center}
\begin{tabular}{l|c|c|cc}
\hline
View Type & NDS$\uparrow$  & mAP$\uparrow$  & mATE$\downarrow$  \\ \hline \hline

Bird's Eye View                    & 56.7 & 47.9 & 0.58  \\ 
Frustum View(u,v=1,d)           & \textbf{58.0} & \textbf{49.4} & \textbf{0.54}  \\
Improvement                & \textbf{+1.3} & \textbf{+1.5} & \textbf{-0.04} \\ 
\hline
\end{tabular}
\end{center}
\label{table5}
\vspace{-25pt}
\end{table}
(ROSCA and RCSCA), as shown in Table~\ref{table4}. (A) represents the baseline, BEVFormer-S-DFA3D~\cite{dfa3d}. When we compare (A) with (B), it indicates that solving the same-ray-same-feature problem using only depth distribution obtained from images is challenging due to inaccurate depth information, and we can enhance it by utilizing radar occupancy to obtain more accurate depth information. Moreover, comparing (B) with (D), incorporating radar-context information via spatial cross attention leads to improvement. We find that the information in radar context (RCS, Doppler velocity) helps us understand the 3D scene. Lastly, we conjecture that the greater improvement from (B) to (D) compared to (C) to (D) is because RCSCA has already learned the sampling offset in deformable attention~\cite{deformable} to refine the BEV features containing depth-ambiguous image characteristics obtained earlier~\cite{dfa3d}.

\subsubsection{Analysis of radar context feature view} Table~\ref{table5} presents the performance results of spatial cross-attention in the proposed RCSCA module, comparing the utilization of radar context feature maps in the frustum view versus the BEV. Here, we do not use ROSCA to isolate the effects of RCSCA based on different views. The results show that utilizing radar context features from the frustum view yields better performance than from the BEV space. This improvement is attributed to CRAB, which fetches image features from the same space in ROSCA, ensuring consistency in obtaining corresponding features in RCSCA, eliminating view discrepancies, and thereby enabling the network to learn more consistently and efficiently.\begin{table}[hbt!]
\vspace{0pt}
\caption
{
    Analysis of different lighting and weather conditions.  \textbf{Bold} and \underline{underline} denote the best value and second best results.
}
\begin{center}
\begin{tabular}{l|c||c c c c}
\hline
  Method    & Modality & Sunny & Rainy & Day  & Night \\ \hline \hline
BEVFormer-DFA3D\cite{dfa3d} & C        & 50.5  & 51.2  & 48.0 & 27.8  \\ \hline 
CenterPoint-P\cite{centerpoint} & L        & 59.7  & 59.2  & 60.1 & 34.7  \\ 
CenterPoint-V\cite{centerpoint} & L        & \textbf{65.1}  & \underline{64.8}  & \textbf{65.5} & \underline{39.3}  \\ \hline 
REDFormer\cite{redformer}    & C\&R      & 48.2  & 50.9  & 48.9 & 28.1  \\ 
TransCAR\cite{transcar}    & C\&R      & 46.8  & 49.6 & 47.4 & 24.6  \\ 
CRN\cite{crn}    & C\&R      & 60.9  & 62.8  & 61.2 & 35.5  \\ 
\textbf{CRAB}     & C\&R      & \underline{61.8}  & \textbf{65.3}  & \underline{62.4} & \textbf{40.0}  \\ \hline
\end{tabular}
\label{table6}
\end{center}
\vspace{-20pt}
\end{table}

\subsubsection{Lighting and Weather Conditions}We measure NDS on the validation set across various weather and lighting conditions, as shown in Table~\ref{table6}. Thanks to the radar's robustness in adverse weather, CRAB consistently outperforms the camera-only baseline~\cite{dfa3d} in all scenarios. Compared to other camera-radar fusion models~\cite{crn,transcar,redformer}, our approach achieves superior performance. This demonstrates the effectiveness of our fusion method in leveraging the strengths of different modalities. Additionally, the comparison between CenterPoint-P and CRAB in rainy and night conditions further highlights the radar's robustness in challenging environments.
\section{CONCLUSIONS}
In this paper, we propose CRAB, a camera-radar fusion 3D object detection and segmentation network, to better mitigate the depth ambiguity inherent in backward projection. Our approach utilizes radar occupancy and depth distribution to enable more accurate depth differentiation. Subsequently, we incorporate context information from radar, which enhances 3D perception. The quantitative and qualitative experiments provide evidence that our proposed method alleviates depth ambiguity. For future work, we plan to apply our method to other backward projection-based methods~\cite{detr3d,bevformerv2,polarformer} for generalization and extend to occupancy prediction~\cite{surroudocc}.

\section{ACKNOWLEDGMENT}
This work was supported by Institute of Information \& communications Technology Planning \& Evaluation (IITP) and the National Research Foundation of Korea(NRF) funded by the Korea government(MSIT) under Grants RS-2023-00236245 and 2022R1A2C200494414.
\bibliographystyle{IEEEtran}
\bibliography{Section/ref}

\end{document}